\documentclass{article}
\usepackage{arxiv}

\usepackage{subfig}
\usepackage{hyperref}
\usepackage{color}
\usepackage[roman]{parnotes}

\usepackage{tabularx} 
\usepackage{float}
\usepackage{amsmath,amssymb,amsfonts}
\usepackage{textcomp}
\usepackage{algorithmic}
\usepackage{siunitx}
\usepackage{algorithmic}
\usepackage{xcolor}

\usepackage{hyperref}

\usepackage{graphicx}
\usepackage{subfig}
\usepackage{multirow}
\usepackage{dblfloatfix}

\begin{document}
 
\title{MULTICAST: MULTI Confirmation-level Alarm SysTem based on CNN and LSTM to mitigate false alarms for handgun detection in video-surveillance}

\author{
R. Olmos \\
Dpt. of Computer Science and Artificial Intelligence\\   Andalusian Research Institute in Data Science and\\ Computational Intelligence, DaSCI \\ University of Granada\\ 18071 
  Spain\\ 
 \And
  S. Tabik\thanks{Corresponding author} \\
Dpt. of Computer Science and Artificial Intelligence\\ Andalusian Research Institute\\ in Data Science and\\ Computational Intelligence, DaSCI \\ University of Granada, 18071  Spain\\ 
  \texttt{siham@ugr.es} \\
  \And
  Francisco P\'erez-Hern\'andez \\
Dpt. of Computer Science and Artificial Intelligence\\ Andalusian Research Institute\\ in Data Science and\\ Computational Intelligence, DaSCI \\ University of Granada, 18071  Spain\\ 
  \And
 Alberto Lamas \\
Dpt. of Computer Science and Artificial Intelligence\\ Andalusian Research Institute\\ in Data Science and\\ Computational Intelligence, DaSCI \\ University of Granada, 18071  Spain\\ 
    \And F. Herrera\\
Dpt. of Computer Science and Artificial Intelligence\\ Andalusian Research Institute\\ in Data Science and\\ Computational Intelligence, DaSCI \\ University of Granada, 18071  Spain\\ 
}

\maketitle

\begin{abstract}

Despite the constant advances in computer vision, integrating modern single-image detectors in real-time handgun alarm systems in video-surveillance is still debatable. Using such detectors  still  implies a high number of false alarms and false negatives. 
In this context, most existent studies select one of the latest single-image detectors and train it on a better dataset or use some pre-processing, post-processing or data-fusion approach to further reduce false alarms. However, none of these works tried to exploit the temporal information present in the videos to mitigate false detections. This paper presents a new system, called MULTI Confirmation-level Alarm SysTem based on Convolutional Neural Networks (CNN) and Long Short Term Memory networks (LSTM) (MULTICAST), that leverages not only the spacial information but also the temporal information existent in the videos for a more reliable handgun detection. MULTICAST consists of three stages, i) a handgun detection stage, ii) a CNN-based spacial confirmation stage and iii) LSTM-based temporal confirmation stage. The temporal confirmation stage uses the positions of the detected handgun in previous instants to predict its trajectory in the next frame. Our experiments show that MULTICAST  reduces by  80\% the number of false alarms  with respect to  Faster R-CNN based-single-image detector, which makes it more useful in providing more effective and rapid security responses.

\end{abstract}

  


\section{Introduction}

The  awareness of potential dangers is crucial for an effective and rapid security response. In most public places, for example in jewelries, banks or train stations, the simple presence of a weapon, such as a handgun, generates a  situation of danger. If a handgun is detected within these areas, a security response should be given as soon as possible. The automatic detection of this kind of objects in video-surveillance can surely help reducing the amount of harm a delinquent can inflict.

The detection of handguns in video-surveillance  is still an open issue as handguns are small,  usually occluded with hands and can be made of reflective metallic materials. The complexity of the detection increases considerably in situations of danger due to sudden and violent movements which together with the large variability in light conditions decreases  the quality of the frames. The later frequently produces   false positives and the mitigation of these false positives is in fact a challenge in these scenarios \cite{zohrevand2019should}.

All the works on weapon detection in videos apply one of the most influential single-image object detection models such as Faster R-CNN \cite{ren2015faster}, SSD \cite{liu2016ssd}, YOLO \cite{redmon2016you}, Efficientdet~\cite{tan2020efficientdet} or NAS-FPN ~\cite{ghiasi2019fpn}. The first studies in this context improved the detection in videos by building new datasets \cite{olmos2018automatic,SALAZARGONZALEZ2020297, warsi2020automatic,lim97deep,jain2020weapon}. The most recent studies applied  pre- or post-processing techniques  to further reducing the errors \cite{mahajan2018detection,vallez2020deep,castillo2019brightness,perez2020object}. While other studies used data fusion to improve the overall performance \cite{truong2020detecting, basit2020localizing,olmos2019binocular, 9032904,ruiz2020handgun}. However, none of these works tried to leverage the temporal information existent in the videos in order to mitigate the detection of false positives  and hence reduce false alarms in handgun detection.

This paper presents MULTICAST,  MULTI Confirmation-level Alarm SysTem based on Convolutional Neural Networks (CNN) and Long Short Term Memory networks (LSTM) (MULTICAST), that leverages not only the spacial information but also the temporal information existent in the videos in order to provide a more reliable handgun alarm system. MULTICAST consists of three stages:
\begin{itemize}
    \item A handgun detection stage. This stage analyses the entire input frame searching for a potential handgun.
    \item A CNN-based spacial confirmation stage.  When the first stage finds a possible handgun,  the spacial confirmation stage double checks whether this detection is an actual true positive.
    \item An LSTM-based temporal confirmation stage. When activated, the third stage uses the positions of the detection from previous instants to predict its trajectory in the next frame. This stage is essential for recovering possible false negatives produced by the first stage.

\end{itemize}
Each stage updates an index that we name Confirmation Level (CL) to indicate how confident it is about the presence of a handgun in the frame (for stage 1) or in a region of the frame (for stage 2 and 3).

The main contributions of this paper can be summarized as follows: 

\begin{itemize}
    \item To present MULTICAST (MULTI Confirmation-level Alarm SysTem based on CNN and LSTM), a  reliable handgun detection that reduces substantially   the  number of false alarms. 
    
    \item To provide HandgunTrajector\footnote{HandgunTrajectory database can be found in \url{https://github.com/ari-dasci/OD-MULTICAST}}, a new training dataset for the prediction of  handguns trajectory in videos. This dataset is used to train the temporal confirmation LSTM-based network.

\end{itemize}

Our experimental results on fifteen test-videos shows that MULTICAST based on  Faster R-CNN for handgun detection, EfficientDet for the spacial confirmation and LSTM for the handgun trajectory prediction, reduces the number of false alarms by 80\% with respect to  Faster R-CNN based-single-image detector.


The rest of the paper is organized as follows. Section~2 reviews related works on handgun detection in videos. Section~3 presents the details of MULTICAST. Section~4  experimentally analyzes the capability of MULTICAST as single-image handgun detector and as alarm system and compares it with the most accurate single-image detector in handgun detection. Finally, Section~5 provides conclusions.

\section{Related works}

Most related works on visible weapon detection in RGB  videos select one of the state-of-art single-image detectors and trains it on a new weapon dataset. The most recent studies in this context intend to further reduce the number of false positives (FP) and false negatives (FN) using either pre-processing, post-processing or data-fusion. As far as we know, there does not exist  any study on leveraging the temporal information existent in the videos with the objective to improve the detection of handguns in video-surveillance.


The seminal paper in handgun detection in videos is~\cite{olmos2018automatic}, where the authors trained Faster R-CNN on a new handgun database\footnote{https://github.com/ari-dasci/OD-WeaponDetection/tree/master/Pistol\%20detection}. The obtained model achieves good results on high quality YouTube videos but produces an important number of FP and FN on lower quality videos. Similarly, the authors in  \cite{SALAZARGONZALEZ2020297} built a dataset using images from a Closed-Circuit TeleVision (CCTV) installed at the University of Seville and synthetic data generated by Unity game engine\footnote{https://unity.com/}. They showed that Faster R-CNN based on Feature Pyramid Network and ResNet-50 can be used as weapon detector in quasi real-time CCTV.



Several works addressed reducing the number of FP and FN in handgun detection in realistic video surveillance environments using pre-processing \cite{mahajan2018detection,castillo2019brightness} and post-processing approaches \cite{vallez2020deep,perez2020object}. For example, the authors in ~\cite{castillo2019brightness} presented a pre-processing technique that help improving the quality of the videos in changing illumination environments. They showed that this technique improves  the detection of knives in videos. The experimental results in \cite{vallez2020deep} showed that analyzing the  detected handguns using  an auto-encoder  improves the overall precision of the detection in videos. In ~\cite{perez2020object}, the authors showed that the use of binarization techniques such as OVO and OVA improves the overall performance of  handguns and knives detection. However, this solution implies a high computational cost that makes  real time execution impossible.

Some other works proposed using information fusion to improve the detection of handguns in  videos \cite{olmos2019binocular,truong2020detecting,basit2020localizing,9032904}.  For example, the authors in ~\cite{olmos2019binocular}  fused the binocular information to eliminate an important number of possible FP from the background in the frames. This approach showed a high potential in reducing the  number of FP. However, the required setup  is not available in regular video surveillance environments, i.e., two synchronized symmetric cameras set at a specific distance and orientation angles.
In \cite{9032904}, the authors  used the persons skeletal pose estimate to detect the threat in an image. They designed a multi-stage classification model. Namely, a  first CNN determines whether a person and a handgun are present in an image. If so, a second CNN estimates the pose of the person and finally a feed-forward neural network assesses the threat level based on the joint positions of the persons skeletal pose estimate from the previous stage. The main drawback of this approach is that it does not perform a detection task, it only classifies individual frames. Similarly, in \cite{ruiz2020handgun}, the authors first use the person pose estimate to locate the areas of the hands in the image then analyze the presence of the handgun in those regions. The main limitation of this approach lies the strong dependence between pose estimation and handgun detection,  if the pose estimation fails, the detection of the handgun also fails.





Nevertheless, there does not exist any work on using the temporal information available in videos with the objective to improve the detection of handgun in video-surveillance. The few existing works try to leverage such information  to improve object detection in general \cite{bi2020sta}.  For example,  to achieve a faster object-detection in mobile CPUs,  the authors in ~\cite{liu2018mobile}  combined a fast single-image object detection with convolutional LSTM layers to create an interweaved recurrent-convolutional architecture.   The authors in  ~\cite{lu2017online} used the LSTM network    to improve the association  between objects of different frames and hence improve the detection of objects in videos in general. They  feed the LSTM with the visual features resulting from the SSD detector, the information of locations and predicted classes. This approach allows recovering lost detections and correcting the predicted class in case it changes radically. 

Differently to the aforementioned works, the present work leverage the temporal information  independently on the CNN visual features. We use an LSTM network to predict the trajectory of the handgun based only on its historical positions, bounding box coordinates.

\section{MULTICAST: MULTI Confirmation-level Alarm SysTem based on CNN and LSTM}

\begin{figure}[H]
	\centering
	\includegraphics[width=0.9\textwidth]{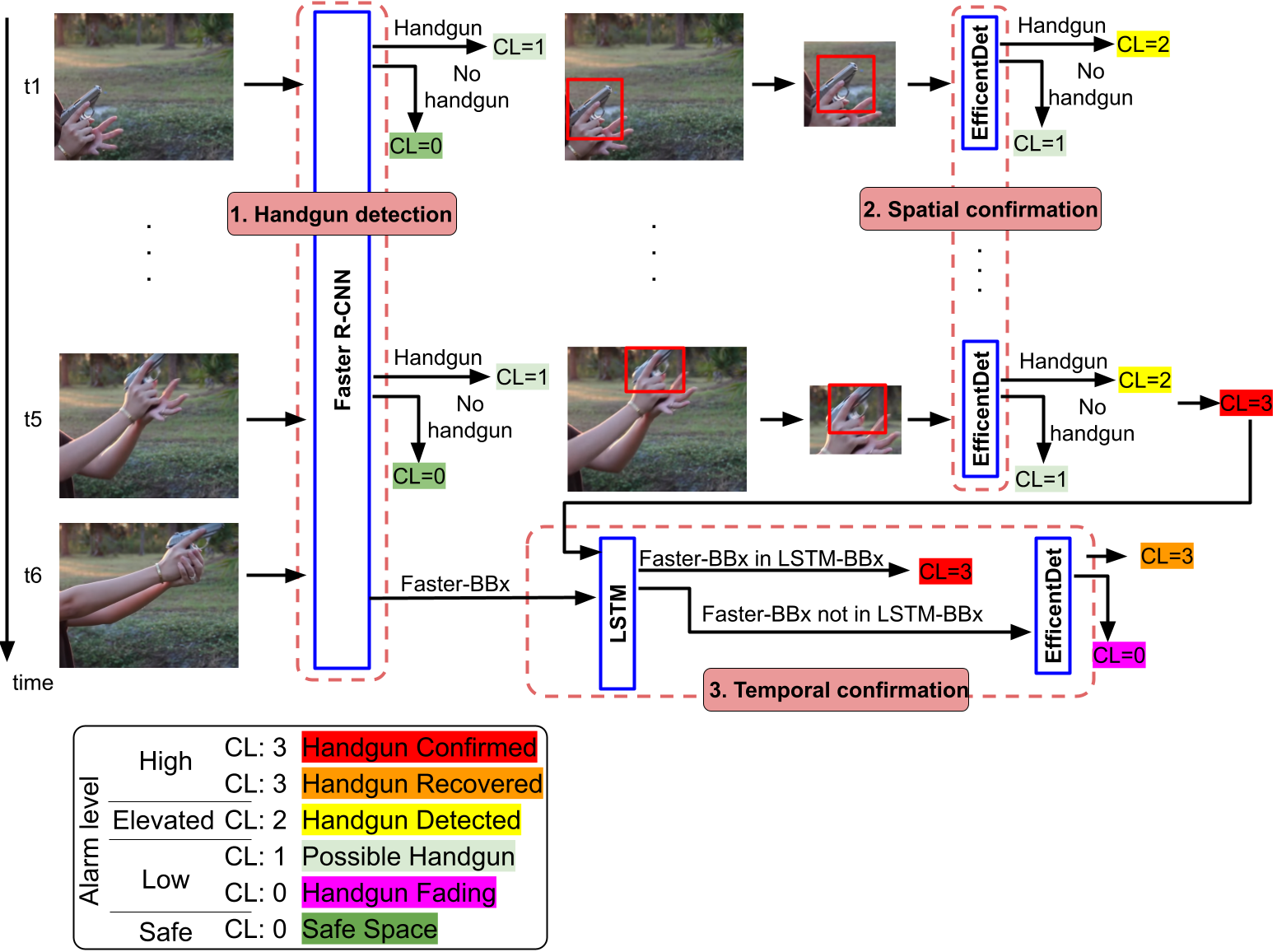}
	\caption{Overview of the proposed MULTI Confirmation-level Alarm SysTem based on CNN and LSTM. MULTICAST includes three stages, handgun detection, spatial confirmation and temporal confirmation.} \label{fig_Multilevel}
\end{figure}

The proposed system is composed of three stages, i) handgun detection, ii)  spatial confirmation and iii) temporal confirmation, as shown in  Figure \ref{fig_Multilevel}. Each stage includes all the necessary mechanisms to guarantee the correct functioning of the whole alarm system in all the situations, as described in the following subsections. Each stage updates an index that we name Confirmation Level (CL) to indicate how confident it is about the presence of a handgun in the frame (for stage 1) or in a region of the frame (for stage 2 and 3).

For a simple communication with the end-user,  the CL value is translated into global alarm level about the presence of danger in the whole frame. We define four alarm levels, safe, low, elevated and high. The correspondence between each CL value and alarm level is shown in  Table \ref{tab:1}. For a more accurate feedback to the end-user, an extra useful information about the state of the handgun detection, i.e., plotted in right lower corner of the frame(s). 

\begin{table}[H]
\centering
\caption{Definition of the the alarm levels and corresponding Confirmation Level (CL) values.}
\label{tab:1}
\resizebox{\textwidth}{!}{
\begin{tabularx}{\textwidth}{ccXX}
\hline
\begin{tabular}[c]{@{}c@{}}Alarm\\ level\end{tabular} &  CL & Meaning & \begin{tabular}[c]{@{}c@{}}Shown message\\ in the frame\end{tabular} \\\hline\hline
\textbf{Safe}  & 0 & No handgun is present in the scene. &  {\it{Safe Space}}\\\hline

\textbf{Low}  & 0,1 & CL is set to 1, when the first stage has detected a possible handgun in at least one frame. CL is decreased to 0 when the temporal confirmation has not recovered a detection \parnote{Notice that the value of CL depends on what happened in  the current frame and also on what happened in the previous instants}. & \it{Handgun Fading}  for CL=0  \it{Possible Handgun} for CL=1 \\\hline

\textbf{Elevated}  & 2 & A detected handgun is confirmed by the spacial confirmation stage in at least one frame. This is considered as a critical situation that requires human verification. & \it{Handgun Detected} \\\hline 

\textbf{High}  & 3& A detected handgun is confirmed by the spacial confirmation in five frames and/or by the temporal confirmation. This  value is kept as 3 when a handgun detection is recovered in at least one of the subsequent five frames after a  false negative by stage 1. This is considered as highly critical situation that requires immediate attention. & \it{Handgun Confirmed}\\\hline 
\end{tabularx}}
\parnotes
\end{table}

\subsection{Stage 1: Handgun detection}

This  is the main stage for the detection of handguns in individual frames. This stage analyzes globally the whole input frame using Faster R-CNN and searches for a possible handgun.  If a possible handgun is detected in a frame, its associated CL is increased to 1 and the area around the detected handgun with an extra padding is cropped and sent for its further analysis to the spacial confirmation stage. When multiple handguns are present in the frame, the same procedure is applied for each detected handgun. The next stages, stage 2 and 3, are activated  for each individual handgun detection.

\subsection{Stage 2: Spatial confirmation}
The spacial confirmation stage is based on a CNN-based detection model such as EffiecientDet. This stage is activated when a handgun is found by the main stage. In particular, the cropped region of each detected handgun by Faster R-CNN  is analyzed by this stage using a different type of detectors, in this study we used EffiecientDet. If the detection is confirmed in at least one frame, CL is increased to level 2. If the detected handgun is confirmed in five consecutive frames, CL is set to 3, the overall alarm level is set to high and the temporal confirmation analysis is activated. 

Notice that this stage analyzes each detected handgun individually.

\subsection{Stage 3: Temporal confirmation}

The temporal confirmation stage is based on an LSTM-network to predict the trajectory, of the confirmed handgun by stage 2, in the next frame.  The predicted trajectory actually refers to  the  bonding box of the area where the handgun could possibly move to in the next frame. Each bounding box is determined by the X and Y coordinates of the left bottom corner and right top corner.

\begin{figure}[H]
	\centering
	\includegraphics[width=0.8\textwidth]{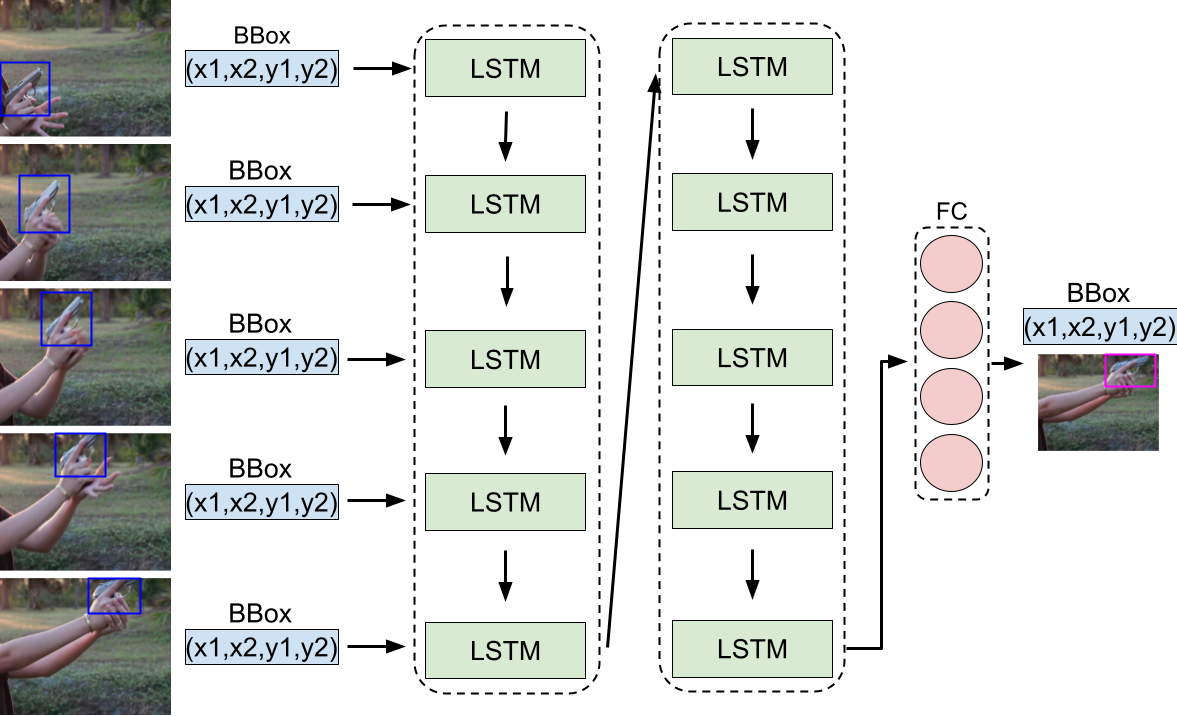}
	\caption{Architecture of the used LSTM-based network used in the temporal confirmation stage.} \label{fig_LSTM}
\end{figure}


The LSTM-based network used in this stage contains two LSTM layers, as it is shown in Figure \ref{fig_LSTM}. Each LSTM cell has 200 units. This architecture has been selected over other architectures due to its simplicity,  low computational cost and robustness. 

For training this network to predict the trajectory of a detected handgun in the next frame, we built a dataset named HandgunTrajectory\footnote{HandgunTrajectory database can be found in \url{https://github.com/ari-dasci/OD-MULTICAST}}. This dataset includes five files obtained from the analysis of five different videos. Each file contains  a list of predicted bounding boxes obtained from Faster R-CNN. Each line of the list represents, the id of the frame, the X,Y coordinates of the left bottom corner and right top corner of the bounding box and the name of the object class.

When activated, the LSTM-network receives from the first stage a list of  bounding box coordinates of the detected handgun in the last five frames. Then, it predicts the bounding box in the next frame. The predicted bounding box is compared with the obtained bounding box from Faster R-CNN after analyzing the sixth frame. If both bounding boxes match, CL value is kept as 3 and this detection will be tagged as {\it Handgun confirmed}. If the  bounding boxes do not match, because probably Faster R-CNN has produced a false positive, a recovering process is performed. Namely, the predicted area by LSTM-net is cropped and analysed by EfficientDet in up to five instants or frames. If EfficientDet finds a handgun in that region of the frame, CL value is kept as 3 and the detection will be tagged as {\it Handgun Recovered}. On the contrary, if EfficientDet does not confirm the presence of handgun in five consecutive instants, CL value is decreased to 0 and the detection is tagged as {\it Handgun Fading}.

Notice that this stage analyzes each detection individually.

\section{Experimental analysis}

In this section, we analyze, compare and evaluate MULTICAST in two different tasks, i) as a single-image handgun detector applied to videos (section \ref{subsection_1}) and ii) as an automatic alarm system (section \ref{subsection_2}). Therefore, we performed two different experiments with different experimental setups. Afterwards, we analyse in more details the false alarms and false negatives mitigation in section \ref{subsection_3} and  section \ref{subsection_4} respectively.

The common considerations for both experiments are as follows. To select the  detection model to be included in stage-1 of MULTICAST and also to be used as baseline for comparison purposes, we performed an exhaustive experimental analysis of different single-image detectors and feature extractors, e.g., SSD, EfficienDet and Faster R-CNN. We found that R-CNN based on ResNet 101  provides the best results in handgun detection and hence, we used Faster R-CNN (ResNet 101) in all the experiments described bellow.

We also used the  dataset\footnote{https://github.com/ari-dasci/OD-WeaponDetection/tree/master/Pistol\%20detection} provided by \cite{olmos2018automatic} for training the detectors included in stage 1 and stage 2 of MULTICAST. 

To better differentiate between the predictions of the different stages in MULTICAST in the frames of the  videos,  the bounding boxes produced by stage 1 are highlighted in blue color,  the bounding boxes produced by stage 2 are highlighted in green color and the predicted bonding boxes by the LSTM stage are highlighted using the  purple color.

\subsection{MULTICAST vs. Faster R-CNN in the task of single-image handgun detection}
\label{subsection_1}
In this analysis, we evaluate and compare MULTICAST with Faster R-CNN as single-image handgun detectors on five test videos. To ease the comparison, only the correctly detected handguns by MULTICAST with CL=2 and CL=3 are considered as True Positives.

\textbf{Test set videos:} To check the behaviour of MULTICAST as handgun detector at the frame level, we selected  five  videos 
fragments from different YouTube videos  showing guns in different situations. Each video presents  different  detection challenges as  described below:

\begin{itemize}

\item Vid1 \cite{TFBTV_2016}:  This video shows a person shooting quickly in very different angles in an outdoor environment. 

\item Vid2 \cite{ActiveSelfProtection_2017}: This video shows an actual  robbery with handgun in a restaurant indoor. Although the illumination is good, the quality of the video is low.

\item Vid3 \cite{CampodeDemolicion_2017}: This video shows at least three persons in each scene.  It is an outdoor video with good quality but with  deficient illumination. 

\item Vid4 \cite{LACORRENTINAFM_2018}: This video shows an actual jewelry  robbery with handgun. The illumination of this video is not uniform and there are four persons in the scene.

\item Vid5 \cite{TheWarpZone_2016}: This video shows  a fake robbery with three persons,  poor illumination and with  multiple camera angles. The main challenge of this video is that the camera angles change constantly in the video, which constantly interrupts the temporal continuity of the video. This is  clearly an unsuitable video for MULTICAST.

\end{itemize}

\textbf{Evaluation metrics}: To evaluate the performance of MULTICAST as handgun detector at the frame level, we used four metrics, {\it
accuracy}, {\it precision}, {\it recall} (also known
as sensitivity), and {\it F1 measure}, which evaluates the balance between {\it
precision} and {\it recall}. Where

$$accuracy=\frac{TP+TN}{Total~images},$$
$$precision=\frac{TP}{TP+FP},$$
$$recall=\frac{TP}{TP+FN},$$ and
$$F1~measure=2\times\frac{precision \times recall}{precision
+recall}$$

Where, True Positives (TP) refers to the number of handguns correctly
detected in the frames of the input video.  In the case when a given
frame has two visible pistols, the detection model must produce two
bounding boxes that will be considered as two TP only if each
bounding-box has an intersection over union  larger than
$70\%$. False Positives (FP) refer to the number of bounding boxes
produced by the detection model in which there is no pistol or a
tiny part of the pistol. If a pistol is detected more than once, the
number of redundant bounding boxes will be considered as FP, since
we have already applied the Non-maximum suppression method to unify
redundant detection of the same object. True Negatives (TN) refers
to the total number of frames where there is neither visible pistols
nor false positives. False negatives (FN) refers to the number of
visible pistols that are not detected by the object detection
algorithm.

For this Setup, it was not possible to evaluate the mAP or the conventional object detection AP as  MULTICAST relays on a lower  confidence threshold, that is $0.1$ for the first stage detector and $0.3$ for the second stage detector. So a simple AP is calculated as the area of the rectangle caused by the fixed threshold of the experiment as:

$$AP={precision}*{recall}$$

\textbf{Results and analysis}: We evaluated two  MULTICAST versions. The first, MULTICAST(FSL) is based on Faster R-CNN, as handgun detector, SSD as spacial confirmation stage and LSTM as temporal confirmation stage. The second MULTICAST(FEL)  is based on Faster R-CNN, as handgun detector, EfficientDet as spacial confirmation stage and LSTM as temporal confirmation stage.

The analyzed videos  with  MULTICAST(FSL), MULTICAST(FEL) and Faster R-CMM are available through\footnote{https://www.youtube.com/playlist?list=PLiMogcXRP-Fokc\_FcZbXJnbGQLtAyKflh}. The performance of MULTICAST(FSL), MULTICAST(FEL) and Faster R-CNN(Resnet 101) on the test-videos are shown in Table \ref{tab:my-table2}.

\begin{table}[H]
\caption{Performance comparison between MULTICAST(FSL) (Faster R-CNN, SSD, LSTM), MULTICAST(FEL) (Faster R-CNN, EfficientDet, LSTM) and Faster R-CNN (ResNet 101) used as single-image handgun detectors in five test videos.}
\label{tab:my-table2}
\resizebox{\textwidth}{!}{%
\begin{tabular}{l||llllllllll}
\hline
       &                  & TP   & FP  & TN   & FN           & Accuracy         & Recall           & Precision        & F1              & AP \\             
       &    &   &   &   &  &(0-100\%) & (0-100\%)  & (0-100\%) & (0-1)     & (0-1) \\ \hline\hline
Vid1 & Faster R-CNN  & 1665 & 77  & 378  & 246     & 87.76\%          & 87.13\%             & \textbf{95.58\%}          & 0.91 & 0.83          \\
       & MULTICAST(FSL) & 1707 & 110 & 357  & 204   & {88.66\%}        & {89.32\%} & 93.95\% & {0.92} & {0.84} \\
       & MULTICAST(FEL) & \textbf{1782} & 124 & 350  & 129   & \textbf{91.58\%} & \textbf{93.25\%}   & 93.49\%          & \textbf{0.93} & \textbf{0.87} \\ \hline

Vid2 & Faster R-CNN  & 125  & 174 & 426  & {188}   & {69.13\%}        & 39.94\%          & {41.81\%} &  0.41 & 0.17          \\
       & MULTICAST (FSL)& 121  & 39  & 474  & {192} & {74.65\%}        & 38.66\%          & \textbf{75.63\%} & {0.51} & \textbf{0.29} \\
       & MULTICAST(FEL) & \textbf{154}  & 127 & 445  & 159   & \textbf{75.16\%} & \textbf{49.20\%} & {54.80\%} & \textbf{0.52} & {0.27} \\ \hline
       
Vid3 & Faster R-CNN  & 1610 & 131 & 224  & 403     & 78.78\%          & 79.98\%          & 92.48\%          & 0.86          & 0.74          \\
     & MULTICAST(FSL) & 1703 & 62  & 286  & 310     & {85.44\%}        & {84.60\%}        & \textbf{96.49\%} & {0.90}        & {0.82} \\
     & MULTICAST(FEL) & \textbf{1813} & 112 & 263  & 200     & \textbf{89.18\%} & \textbf{90.06\%} & {94.18\%}        & \textbf{0.92} & \textbf{0.85} \\ \hline

Vid4 & Faster R-CNN & 20   & 130 & 2178 & 115  & 90.45\%          & 14.81\%          & 13.33\%          & 0.14          & 0.02          \\
       & MULTICAST(FSL) & 15   & 45  & 2250 & 120   & {93.21\%}        & 11.11\%          & {25.00\%}        & {0.15}        & {0.03} \\
       & MULICAS(FEL)  & \textbf{26}   & 33  & 2274 & 109   & \textbf{94.65\%} & \textbf{19.26\%} & \textbf{44.07\%} & \textbf{0.27} & \textbf{0.08} \\ \hline

Vid5 & Faster R-CNN       & 132  & 136 & 632  & 209& 69.20\%          & 38.71\%          & 49.25\%          & 0.43          & 0.19          \\
       & MULTICAST(FSL) & 130  & 126 & 644  & 211   & {70.11\%}        & 38.12\%          & {50.78\%}         & 0.44          & {0.19} \\
       & MULTICAST(FEL) & \textbf{147}  & 61  & 718  & 194   & \textbf{78.35\%} & \textbf{43.11\%} & \textbf{70.67\%} & \textbf{0.54} & \textbf{0.30}\\ \hline
\end{tabular}%
}
\end{table}

As it can seen from  Table \ref{tab:my-table2}, in general,  MULTICAST(FEL)  overcomes Faster R-CNN on all test videos obtaining higher values of  TP, F1-measure and AP. Recall that MULTICAST triggers an alarm only after a spacial and temporal confirmation process. Although the number of FP is higher for MULTICAST(FEL) on some test videos, most of these false detections are  filtered by the the spacial and temporal confirmation stages. 

In addition,  MULTICAST(FEL) based on EfficientDet as spacial confirmation stage achieves better results than using SSD.


\subsection{MULTICAST vs. Faster R-CNN as handgun alarm system in video-surveillance}
\label{subsection_2}

In this section, we analyze the potential of MULTICAST as handgun alarm system. MULTICAST  triggers an alarm   only when the CL value associated to a handgun detection   reaches   CL=3.  Faster R-CNN triggers  an alarm when a handgun  is detected in a frame.



\textbf{Test videos:} We recorded  ten short videos using two conventional home security cameras. These test videos were named as VideoAlarm from VideoAlarm1 to VideoAlarm10\footnote{The resulting videos of this  analysis can be found in \url{https://drive.google.com/drive/folders/1qvvlP4gCjDKZhZJbGSXf3TIb1e1is_Ru?usp=sharing}}. The first five videos show a handgun. The  other five videos do not contain  handgun  but  show different  situations than can cause a false alarm.

\textbf{Evaluation metrics:} For assessing and comparing the potential of MULTICAST as handgun alarm system, we only evaluate whether the system triggers an alarm or not in the presence or not of an actual handgun in the scene. That is, if an actual handgun present in the video, at least in one frame, and the model detects it, the overall evaluation of the video as a TP. Otherwise,  the overall evaluation of the video is a FP.  If there is no handgun in the video and the system detects one, then the overall evaluation of the video is FN. Otherwise, the overall evaluation of the video is TN.

\textbf{Results and analysis:} 
To evaluate the capacity of MULTICAST to be used as alarm system, we consider the presence of a handgun in the entire scene rather in the individual frames. In other words, we evaluate the capacity  to produce an alarm when there is an actual  handgun in the video.

The results of a comparison between MULTICAST and Faster R-CNN  are shown in the Table \ref{tab:r1}. GT to refers to Ground Truth. We consider GT as a binary value to indicate the presence (GT=1) or not ( GT=0) of handgun in  the video.  GT=1 indicates that at least one handgun detection alarm was triggered in a video. GT=0 indicates that even if sparse handgun detections have occurred, the alarm wasn't triggered for a video.

\begin{table}[H]
\centering
\caption{A comparison between MULTICAST and Faster R-CNN as alarm system.  GT is a global measure to  indicate whether a handgun is present in the video. GT=1 indicated that an alarm must be triggered  and GT=0 must not.}
\label{tab:r1}
{%
\begin{tabular}{l|llll}
\hline
             & \multicolumn{1}{c}{GT} & \multicolumn{1}{c}{Faster R-CNN} & \multicolumn{1}{c}{MULTICAST} & \multicolumn{1}{c}{MULTICAST} \\
             & \multicolumn{1}{c}{} & \multicolumn{1}{c}{} & \multicolumn{1}{c}{(FSL)} & \multicolumn{1}{c}{(FEL)} \\ \hline\hline
VideoAlarm1  & 1                            & 1                            & 0                      & 0                      \\
VideoAlarm2  & 1                            & 1                            & 1                      & 1                      \\
VideoAlarm3  & 1                            & 1                            & 1                      & 1                      \\
VideoAlarm4  & 1                            & 1                            & 1                      & 1                      \\
VideoAlarm5  & 1                            & 1                            & 1                      & 1                      \\
VideoAlarm6  & 0                            & 1                            & 1                      & 0                      \\
VideoAlarm7  & 0                            & 1                            & 0                      & 0                      \\
VideoAlarm8  & 0                            & 1                            & 0                      & 0                      \\
VideoAlarm9  & 0                            & 1                            & 0                      & 0                      \\
VideoAlarm10 & 0                            & 0                            & 0                      & 0   \\ \hline                  
\end{tabular}%
}
\end{table}
As it can be seen from Table \ref{tab:r1}, MULTICAST(FEL) and MULTICAST(FSL) trigger respectively 80\% and 60\% less false alarms than Faster R-CNN.  MULTICAST(FEL) fails only in one video, i.e., VideoAlarm1. This improvement can be explained by the fact that MULTICAST(FEL) spacial confirmation with EfficientDet eliminates a large number of occasional false positives as it checks the presence of a handgun in five consecutive frames. 

The analyzed videos  with  MULTICAST(FSL), MULTICAST(FEL) and Faster R-CNN are available through\footnote{https://www.youtube.com/playlist?list=PLiMogcXRP-Fqom8EmM\_Um4Z-h3XmEVhQs}.

For a more global analysis, we summarized the overall accuracy, precision, recall and F1 measure of MULTICAST(FEL), (FSL) and Faster R-CNN, on the ten test videos,  in Table \ref{tab:Alarm}.  Again, these results clearly show that MULTICAST(FEL) provides substantially better accuracy, precision and F1 over Faster R-CNN.  Although, Faster R-CNN provides a higher recall over MULTICAST, it produces a larger number of false alarms, which makes its use  impractical in video-surveillance environments.

\begin{table}[H]
\centering 
\caption{Overall performance  of  MULTICAST and Faster R-CNN as  alarm systems on the ten test videos.}
\label{tab:Alarm}
{%
\begin{tabular}{l|lll}
\hline
\multicolumn{1}{c}{} & \multicolumn{1}{c}{Faster R-CNN} & \multicolumn{1}{c}{MULTICAST} & \multicolumn{1}{c}{MULTICAST} \\
\multicolumn{1}{c}{} & \multicolumn{1}{c}{} & \multicolumn{1}{c}{(FSL)} & \multicolumn{1}{c}{(FEL)} \\
\hline
Accuracy             & 60.00\%                      & 80.00\%                & 90.00\%                \\
Precision            & 55.56\%                      & 80.00\%                & 100.00\%               \\
Recall               & 100.00\%                     & 80.00\%                & 80.00\%                \\
F1                   & 0.71                         & 0.80                   & 0.89    \\ \hline               
\end{tabular}%
}
\end{table}

The results show that MULTICAST triggers much less  false alarms than Faster R-CNN.  In Table \ref{tab:improvAlarm} the percentage of false alarms and true alarms, and how much improvement represents is shown. 

\begin{table}[H]
\centering
\caption{Improvement (in \%) of MULTICAST  over Faster R-CNN as alarm system.}
\label{tab:improvAlarm}
{%
\begin{tabular}{l|ll}
\hline
                 & \#False Alarm & Improvement vs Faster R-CNN \\ \hline
Faster R-CNN        & 80\% & -             \\
MULTICAST(FSL) & 20\%  &  60\%           \\
MULTICAST(FEL) & 0\%   &    80\%        \\ \hline
\end{tabular}%
}
\end{table}

\subsection{Analysis of false positives eliminated by MULTICAST}
\label{subsection_3}

\begin{figure}[H]%
  \centering
  \subfloat[FP (blue BBx.) committed by stage 1 then corrected and confirmed by the spacial confirmation (green BBx.) ]{\includegraphics[width=0.45\textwidth]{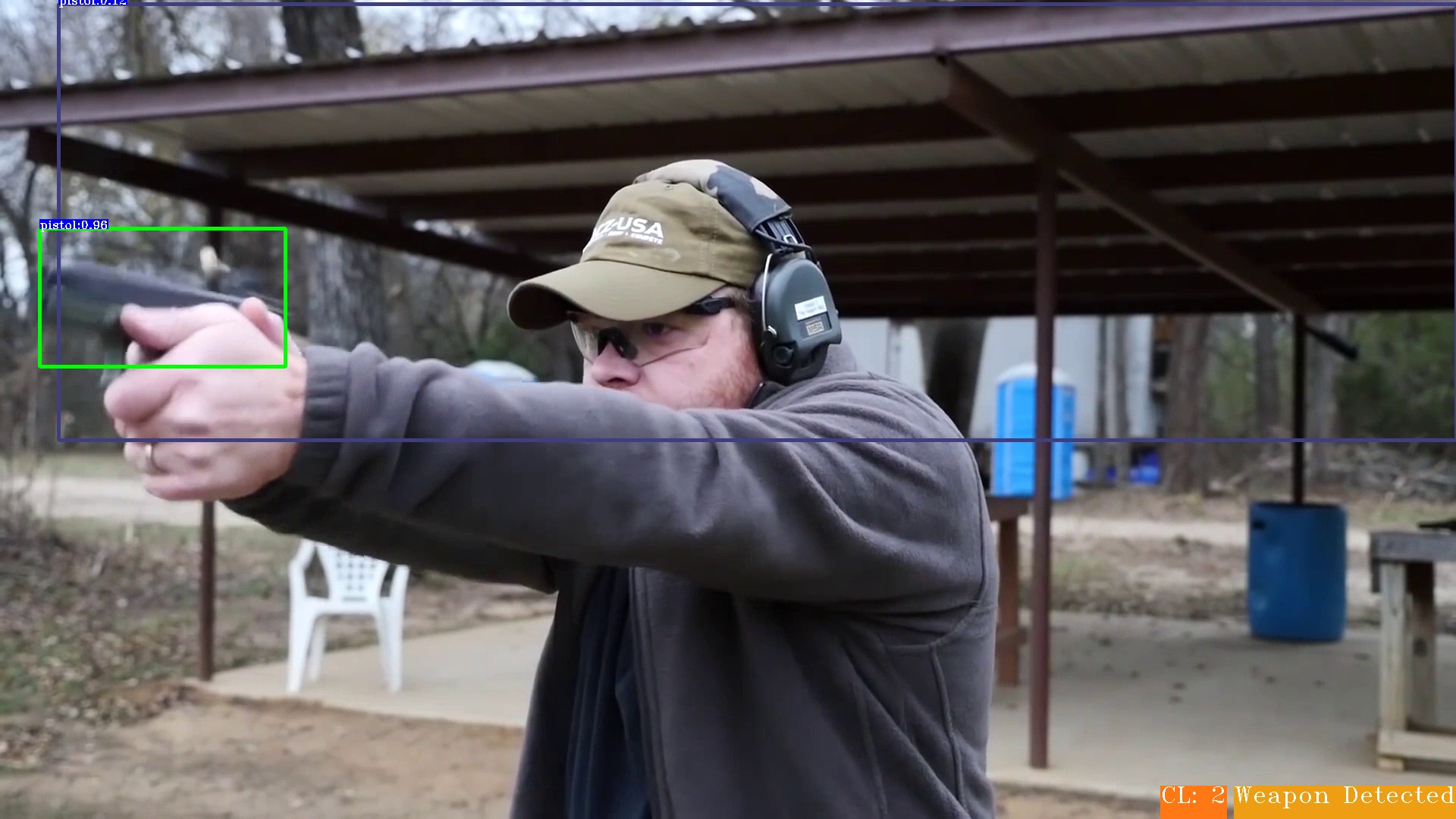} }
    \subfloat[Intermittent type of FP (blue BBx.)  discarded by the spacial confirmation stage]{\includegraphics[width=0.45\textwidth]{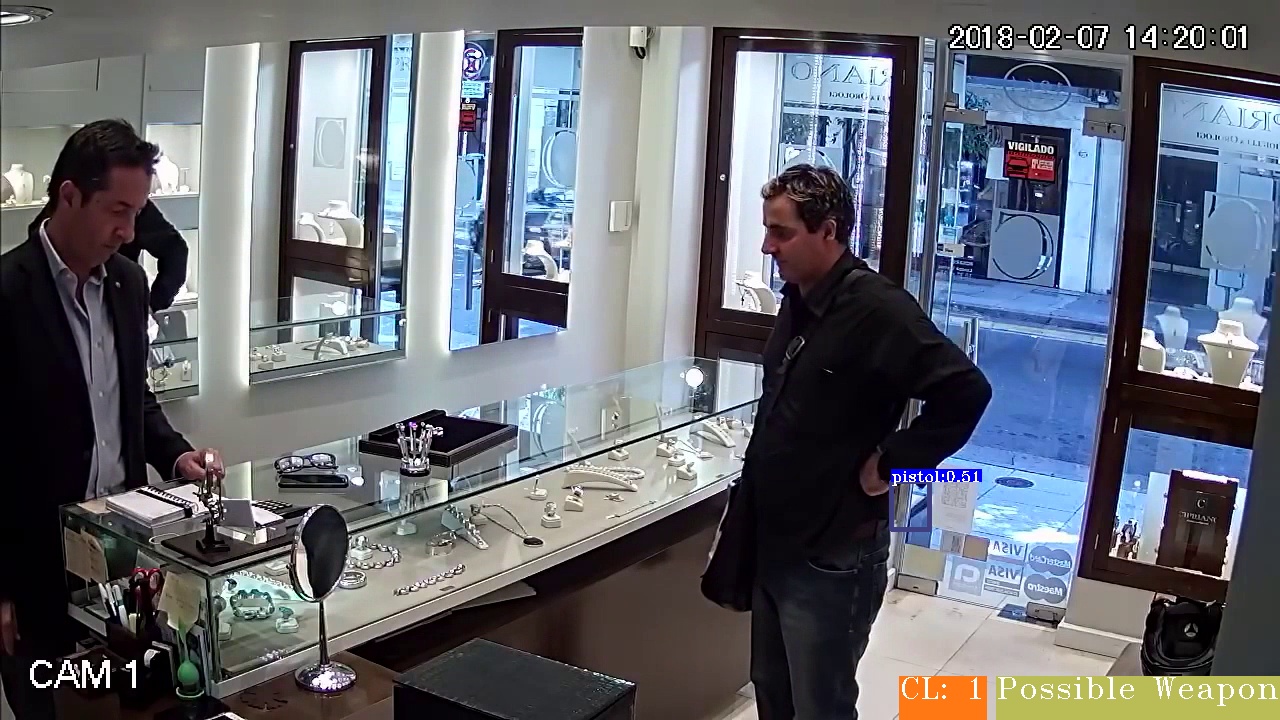}}
   \qquad
  \subfloat[Intermittent type of FP (blue BBx.) discarded by the spacial confirmation stage]{\includegraphics[width=0.45\textwidth]{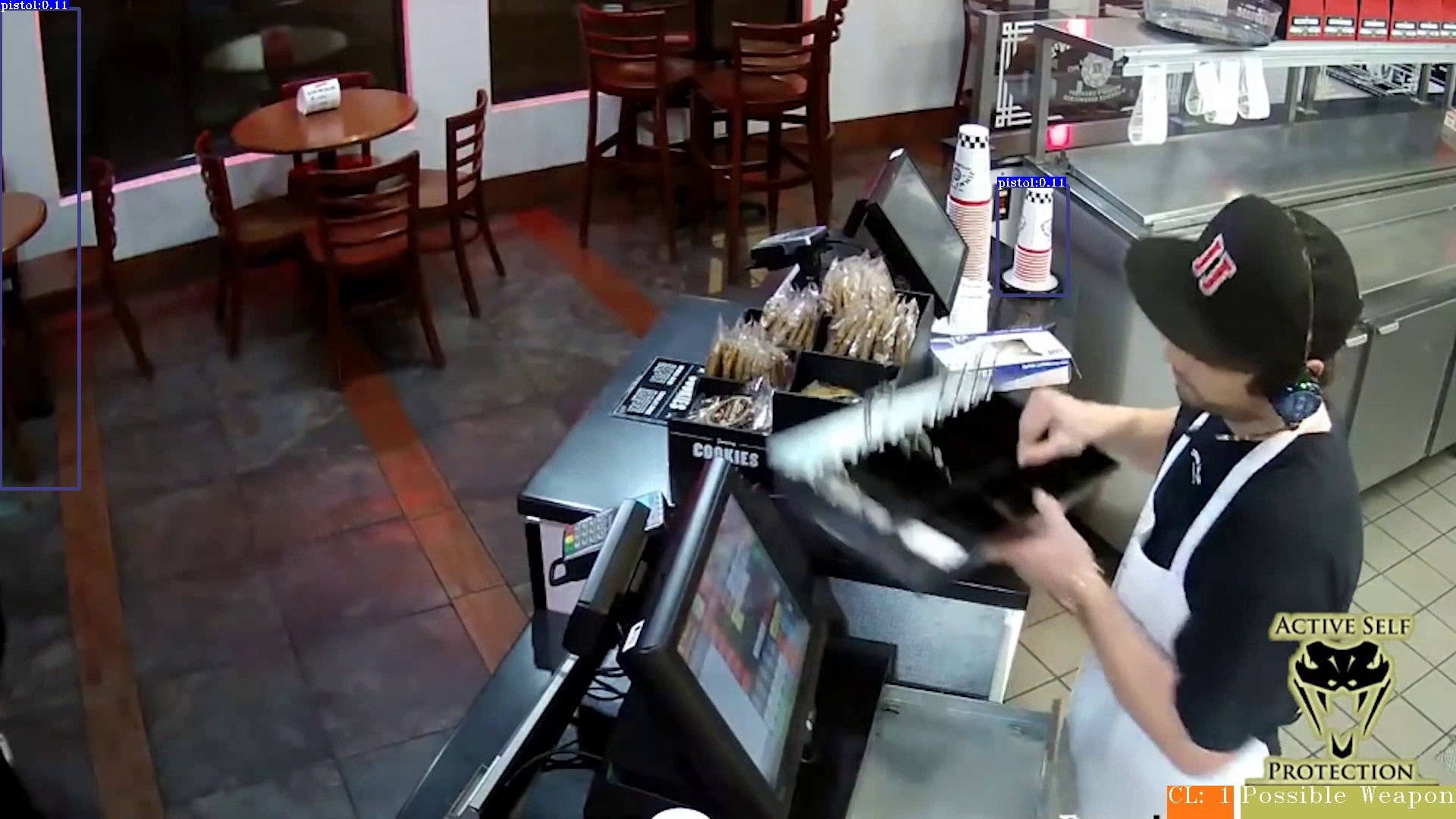}}
\subfloat[Intermittent type of FP (blue BBx. on the left) discarded by the spacial confirmation stage ]{\includegraphics[width=0.45\textwidth]{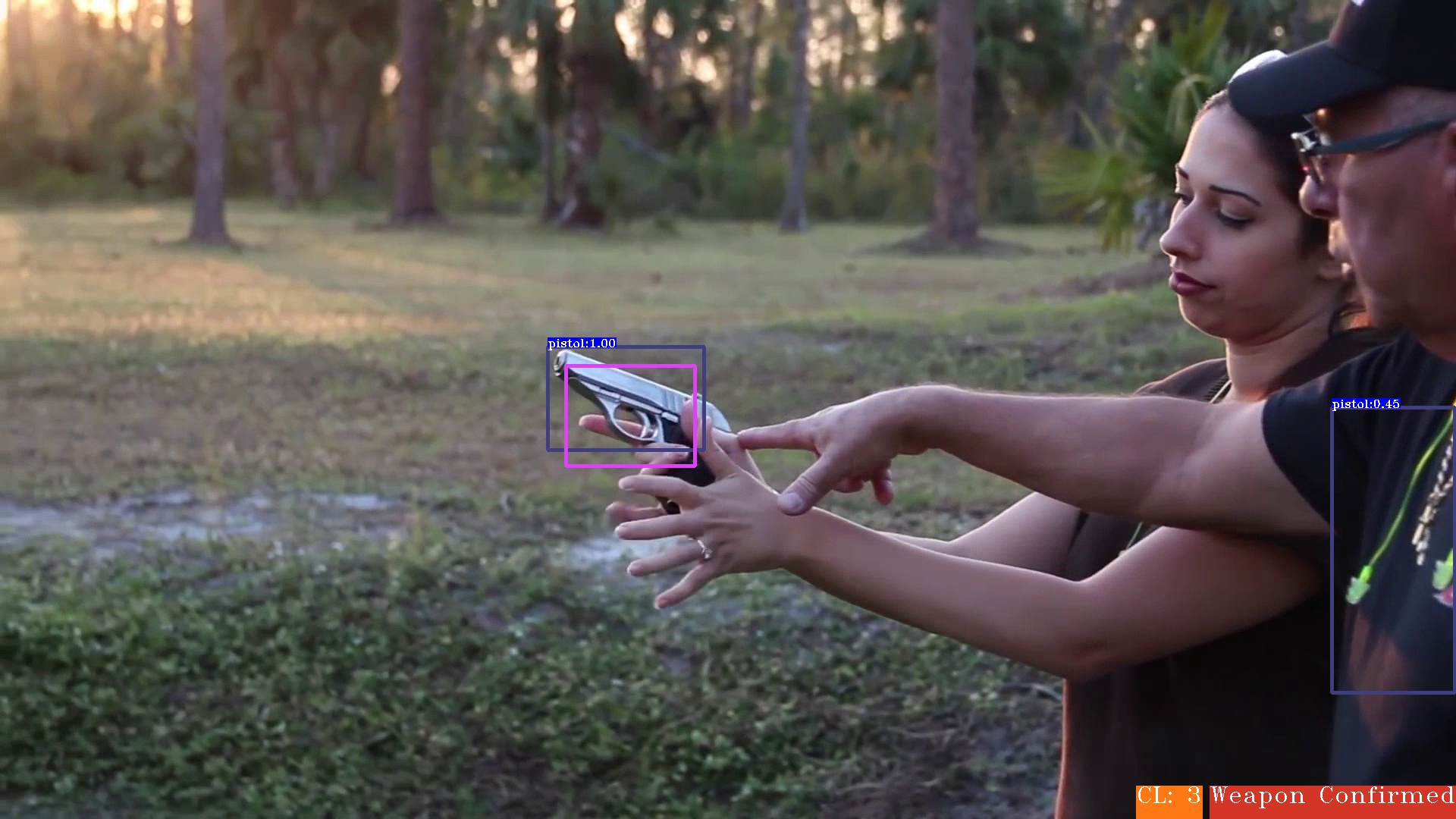}}
 \qquad
 \subfloat[Persistent type of FP (blue BBx.) discarded by the spacial confirmation stage]{\includegraphics[width=0.45\textwidth]{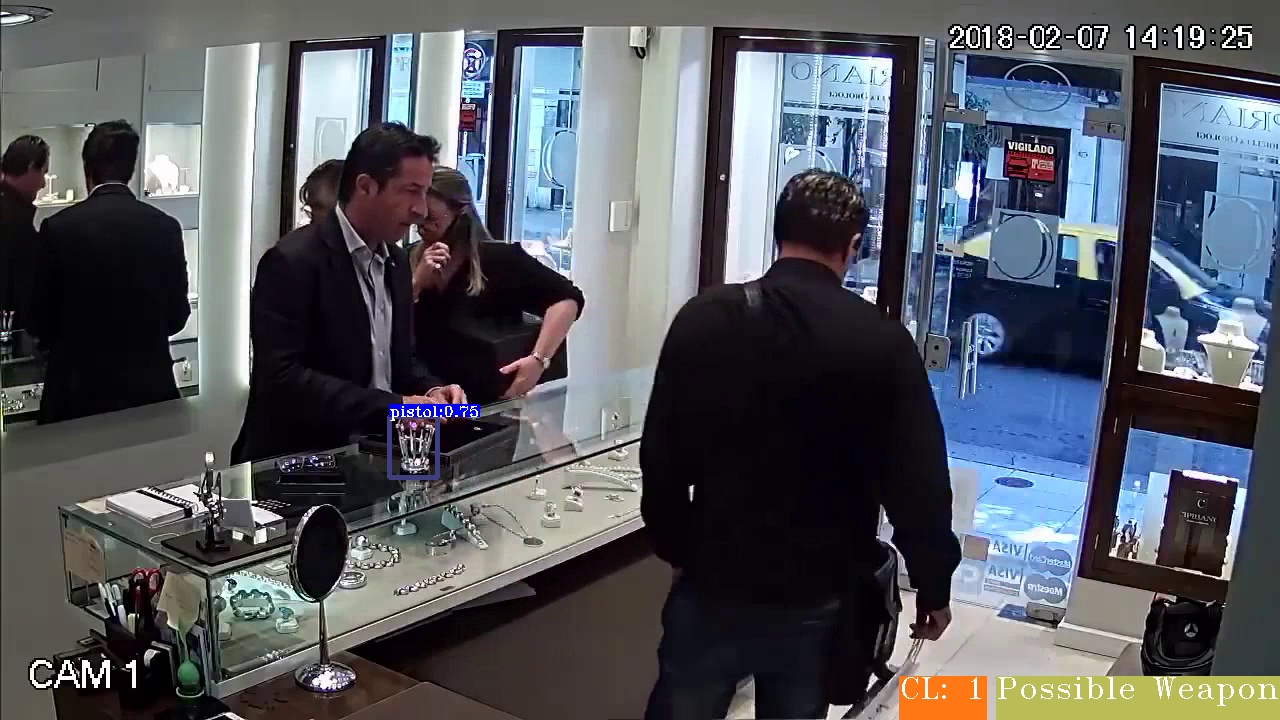}}
 \subfloat[Persistent type of FP (blue BBx.) discarded by the spacial confirmation stage ]{\includegraphics[width=0.45\textwidth]{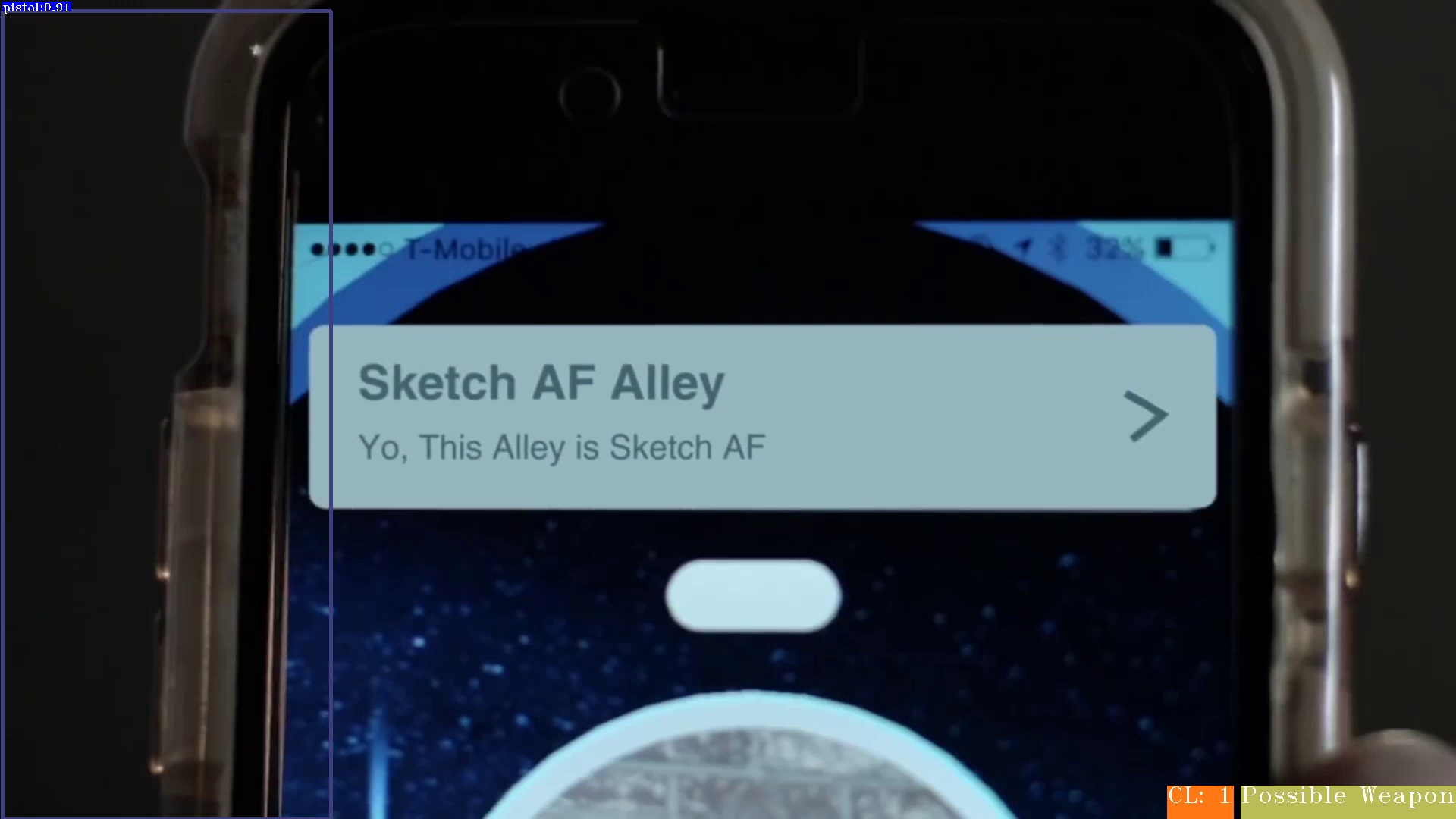}}
 \qquad
\caption{Examples of FP corrected/discarded by MULTICAST. Frames extracted from Videos \cite{CampodeDemolicion_2017},  \cite{LACORRENTINAFM_2018}, \cite{ActiveSelfProtection_2017}, \cite{TFBTV_2016} and \cite{TheWarpZone_2016}.}%
  \label{fig:fpi}%
\end{figure}

After a careful analysis of the errors committed by Faster R-CNN in the first stage and then corrected by MULTICAST, we can distinguish between two types of FP. The first type occurs intermittently showing a varying confidence level, see examples in Figure \ref{fig:fpi}(a), (b), (c) and (d), where the blue color bounding box indicates the false positive committed by Faster R-CNN. In spite of the blur in the handgun area in the frame, MULTICAST was able to correct that erroneous area (delimited by the blue box) by means of the spacial confirmation detector (delimited by the green box).

The second type of FP are persistent FP detections on a particular object such as the glass shown in Figure \ref{fig:fpi}(e) and the cellphone in Figure \ref{fig:fpi}(f). Although the first stage detect these objects as possible handgun, the confirmation stage of MULTICAST confirmed that these detections do not correspond to actual handguns.

In summary, MULTICAST has been able to eliminate most of both types of FP. It has eliminated most of the intermittent FP and reduced substantially persistent ones thanks to its confirmation  mechanisms.

\subsection{Analysis of false negatives recovered by MULTICAST}
\label{subsection_4}

\begin{figure}[h]%
    \centering
    \subfloat[Corrected False Negative 1]{\includegraphics[width=0.45\textwidth]{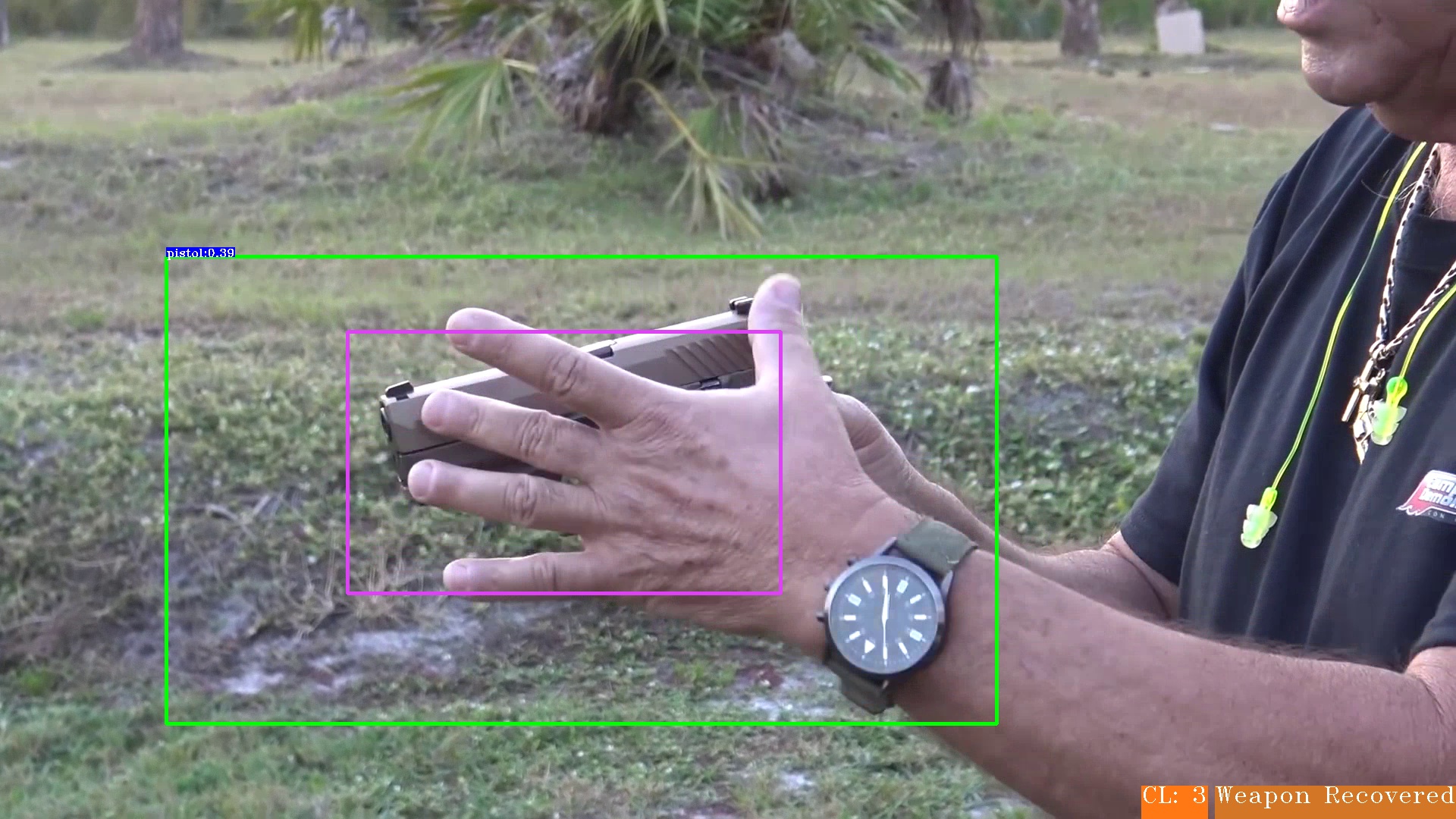}}
    \subfloat[Corrected False Negative 2]{\includegraphics[width=0.45\textwidth]{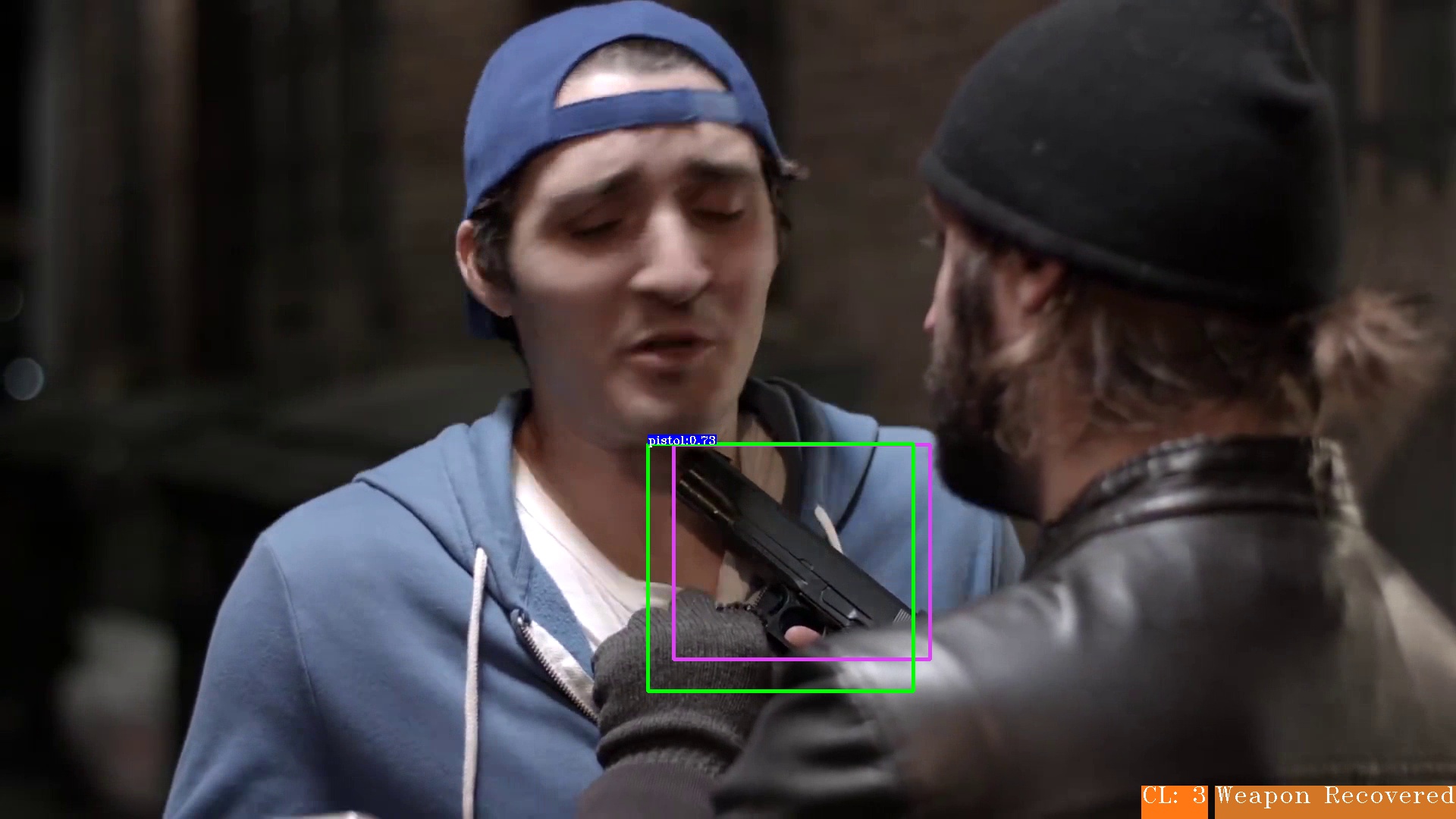}}
    \qquad
    \subfloat[Corrected False Negative 3]{\includegraphics[width=0.45\textwidth]{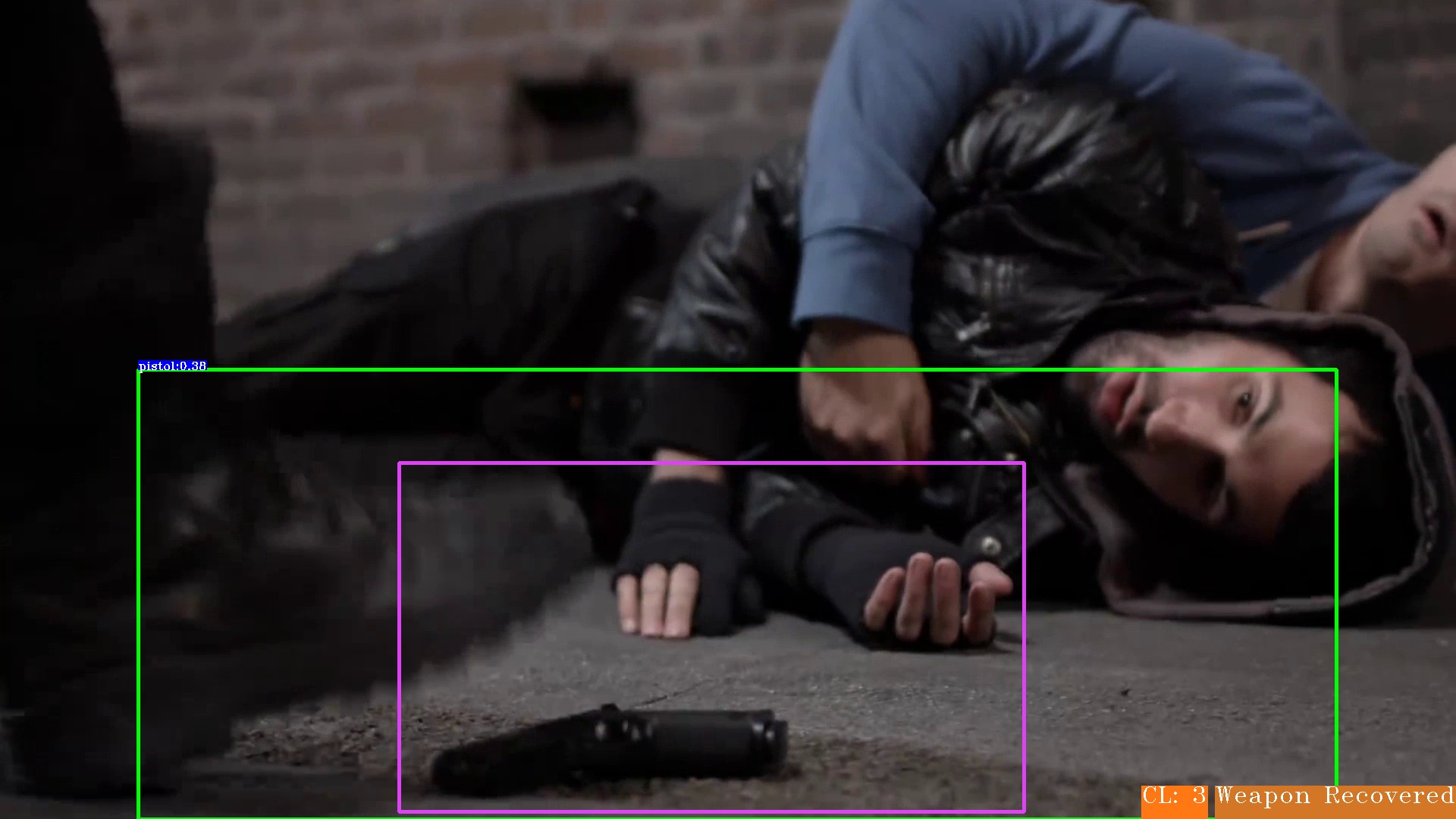}}
    \subfloat[Corrected False Negative 4]{\includegraphics[width=0.45\textwidth]{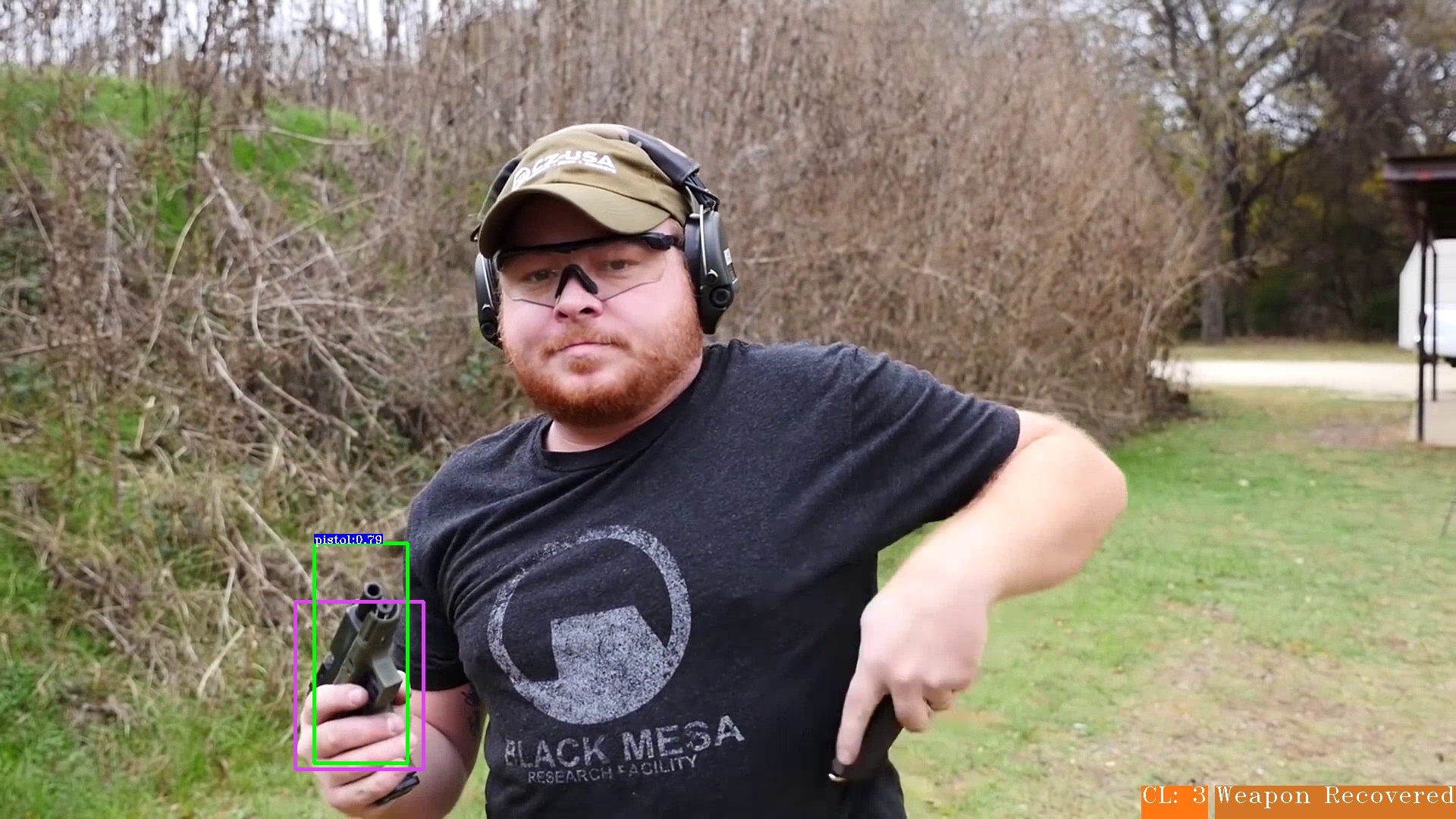}}
    \qquad
    \subfloat[Corrected False Negative 5]{\includegraphics[width=0.45\textwidth]{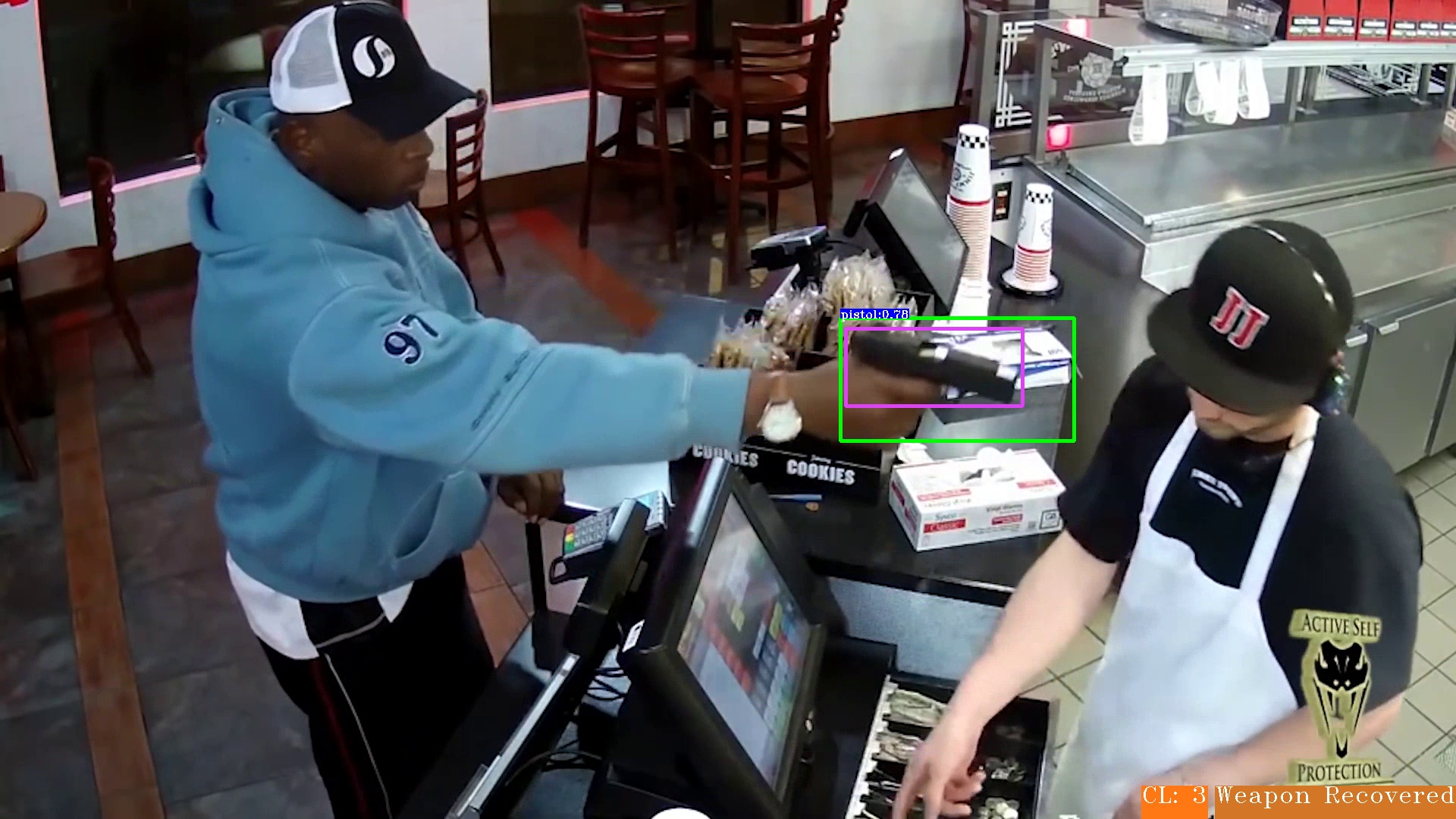}}
    \subfloat[Corrected False Negative 6]{\includegraphics[width=0.45\textwidth]{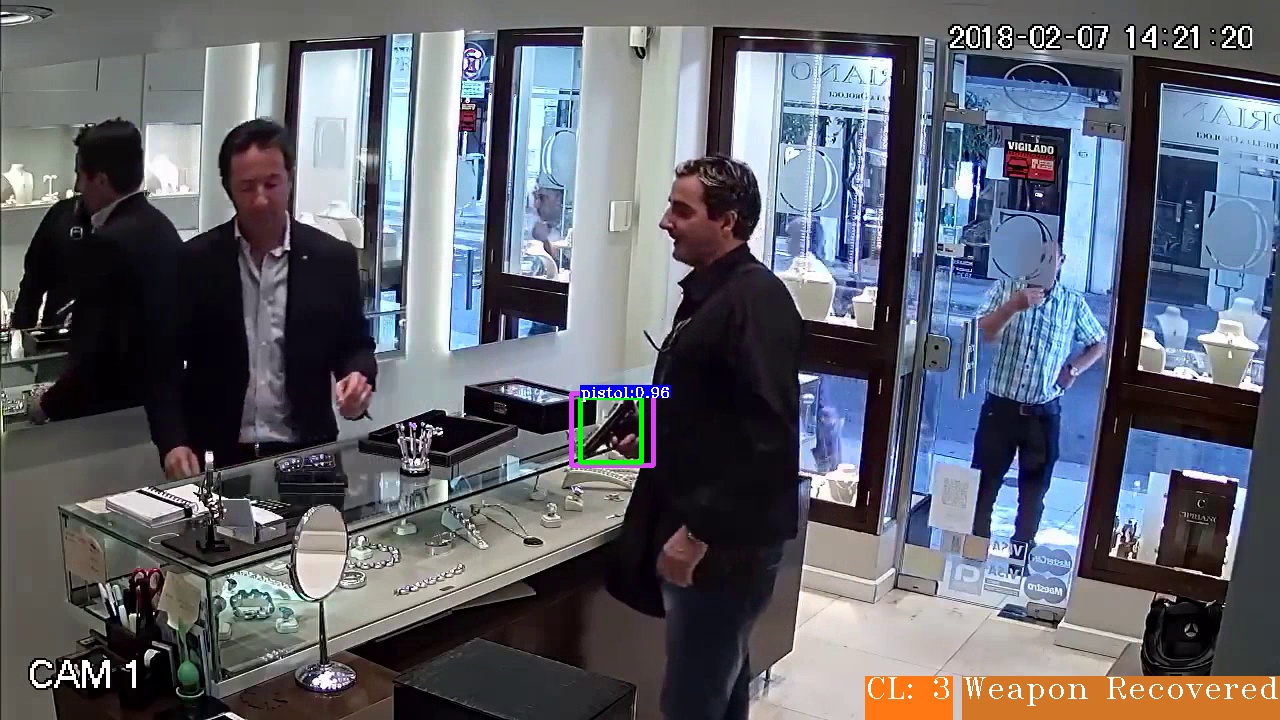}}
    \caption{Examples of False Negatives recovered by MULTICAST. The frames come from videos\cite{CampodeDemolicion_2017},  \cite{LACORRENTINAFM_2018}, \cite{ActiveSelfProtection_2017},  \cite{TFBTV_2016} and \cite{TheWarpZone_2016}.}%
    \label{fig:fni}%
\end{figure}

Faster R-CNN usually produces an important number of FN with a low confidence. Among these FN, we can distinguish between two types. The first type of FN occurs when the detection is difficult because the handgun is far from the camera or due to poor image quality. The second type of FN occurs when  the detector temporally loses
 the handgun due to sudden changes in the light  conditions, position or due to partial occlusion. In these situations MULTICAST is able to recover the detections thanks to the temporal confirmation mechanism as shown in Figure \ref{fig:fni}. The handgun omitted by Faster R-CNN due to partial occlusion in Figure \ref{fig:fni}(a) is recovered by MULTICAST thanks first to the temporal confirmation (purple BBx.) then confirmed by the spacial confirmation stage (green BBx.). The FN in Figure \ref{fig:fni}(b) and (d) were caused by a sudden  change
of the position. The FP in Figure \ref{fig:fni}(c) was caused by a slight change in illumination and position. The FP in Figure \ref{fig:fni}(e) and (f) were due to the bad quality and blur in the frame.

\section{Conclusions}

This work presented MULTICAST,   MULTI  Confirmation-level  Alarm  SysTem  based  on  CNN  and  LSTM,  that  leverages  not  only  the  spacial  information  but also the temporal information existent in the videos for a more reliable handgun detection. MULTICAST allows mitigating the false positives produced by  single-image detection models. Our experimental results showed that thanks to the spacial and temporal confirmation mechanisms, MULTICAST is able to eliminate a large number of FP and FN and hence reduces the false alarms to 80\% with respect to Faster R-CNN.

MULTICAST can help improving the response time and effectiveness in detecting threats in different places by transforming the participation of a human in surveillance into high level  supervision.

As future work, we are working on adapting MULTICAST  to
 Closed-Circuit TeleVision (CCTV) and edge computing devices.

\section*{Acknowledgments}
This work was partially supported by the Spanish Ministry of Science
and Technology under the project:
TIN2017-89517-P and A-TIC-458-UGR18 (DeepL-ISCO). Siham Tabik was supported by the Ramon y Cajal
Programme (RYC-2015-18136).

\bibliographystyle{plain}



\end{document}